\documentclass[runningheads]{llncs}
\usepackage[utf8]{inputenc} 
\usepackage[T1]{fontenc}    
\usepackage{hyperref}       
\usepackage{url}            
\usepackage{booktabs}       
\usepackage{amsfonts}       
\usepackage{nicefrac}       
\usepackage{microtype}      
\usepackage[table]{xcolor}         
\usepackage{graphicx}
\usepackage[super]{nth}
\usepackage{diagbox}

\begin{document}
\title{SkinDistilViT: Lightweight Vision Transformer for Skin Lesion Classification}
%


\author{Vlad-Constantin Lungu-Stan\textsuperscript{\rm 1}  \and Dumitru-Clementin Cercel\textsuperscript{\rm 1} \and Florin Pop\textsuperscript{\rm 1,\rm2}}

\authorrunning{V.-C. Lungu-Stan et al.}


\institute{
    \textsuperscript{\rm 1}Faculty of Automatic Control and Computers, University Politehnica of Bucharest \\
    \textsuperscript{\rm 2}National Institute for Research and Development in Informatics - ICI Bucharest, Romania \\
\email{vlad.lungu@stud.acs.upb.ro, 
\{dumitru.cercel, florin.pop\}@upb.ro}
}

\maketitle              

\begin{abstract}
Skin cancer is a treatable disease if discovered early. We provide a production-specific solution to the skin cancer classification problem that matches human performance in melanoma identification by training a vision transformer on melanoma medical images annotated by experts. Since inference cost, both time and memory wise is important in practice, we employ knowledge distillation to obtain a model that retains 98.33\% of the teacher's balanced multi-class accuracy, at a fraction of the cost. Memory-wise, our model is 49.60\% smaller than the teacher. Time-wise, our solution is 69.25\% faster on GPU and 97.96\% faster on CPU. By adding classification heads at each level of the transformer and employing a cascading distillation process, we improve the balanced multi-class accuracy of the base model by 2.1\%, while creating a range of models of various sizes but comparable performance. We provide the code at \url{https://github.com/Longman-Stan/SkinDistilVit}.

\end{abstract}

\section{Introduction}

\par Skin cancer classification is a crucial problem because health complications can be avoided through early detection and treatment. Deep learning can shine here because both medics and machine learning solutions base their decision on the same information, namely medical images. Since this is important for all humankind, no matter the available computing power, this paper proposes a lightweight, production-ready algorithm that classifies eight types of skin lesions. The algorithm not only provides high performance, but it is also inexpensive to run.
    
\par Since 2017, the mechanism that has revolutionized natural language processing, attention \cite{AttentionIsAllYouNeed}, has shown its prowess for image processing with the vision transformer (ViT) \cite{VisualTransformer}. We opt for an attention-based model because of its versatility and performance. 
However, a problem with the transformer \cite{AttentionIsAllYouNeed} models is their size. Therefore, we use the knowledge distillation technique \cite{hinton_distil} to obtain great performance with a smaller model. We also compare the ViT to convolutional neural networks (CNNs) \cite{kim-2014-convolutional}, the traditional solution for image processing tasks.

\par The difficulty of gathering medical data leads to small datasets being publicly available. Training transformers require huge amounts of data; thus, training one for our melanoma classification task requires extra consideration. Luckily, there are works that train transformer models with considerable data and whose weights are publicly available \cite{rw2019timm}. These models make transfer learning \cite{pan2009survey} possible, enabling the adoption of the ViT for our task.

\par By training a ViT-based solution for the skin lesion classification problem, we match human performance on melanoma identification and obtain a precision of 91.53\% and a recall of 86.73\% for cancer identification in skin lesion images.
Through knowledge distillation, we boost the speed considerably (97.96\%) while reducing the number of parameters almost by half (49.60\%). We also study three ways of producing a series of models of increasing sizes: introducing classification heads after each layer, adding classification heads and forcing their probability distributions to match, and cascading distillation, a technique of gradually distilling away one transformer layer at a time. These techniques boost the base model's performance while creating a range of models that preserve the teacher's performance well.

\par The rest of this paper is organized as follows. In the next section, we present current approaches to our goals. Section 3 details our models, while Section 4 presents the experimental setup. Then, Section 5 describes our results. Lastly, Section 6 concludes the paper.

\section{Related Work}

\par  \textbf{EfficientNets}. Image classification is traditionally solved using CNNs. One prominent set of CNN architectures is the EfficientNet \cite{EfficientNet}. This family of models is the result of a grid search that aims to produce efficient and easy-to-scale models. The authors obtained state-of-the-art (SOTA) performance while drastically reducing the models' size. EfficientNets are the go-to models for competitive image classification tasks on online platforms like Kaggle\footnote{\url{https://www.kaggle.com/}, last visited March 2023.}. These reasons make EfficientNet a good baseline for our task. 

\par \textbf{ISIC 2019 Challenge}. The International Skin Imaging Collaboration (ISIC)\footnote{\url{https://www.isic-archive.com}, last visited March 2023.} is an initiative aimed at alleviating this problem and increasing the performance of melanoma detection systems. 
State of the art for the ISIC 2019 competition \cite{isic2019}
is dominated by ensembles of EfficientNets. The first position \cite{Isic1stPlace} in the contest was obtained by an ensemble of EfficientNet-B0 to B6, while the second place \cite{Isic2ndPlace} was obtained with an ensemble of EfficientNet-B3 to B4. While ensembles are known to behave better than single models, their performance gain of several points is not outstanding, considering the additional computing resources needed for inference. We are interested in high-speed and memory needs, so ensembles are unattractive. 

\par  \textbf{Knowledge Distillation}. Since we aim for a practical approach, even a single model might be too big. 
One solution to tackle this problem is the technique called knowledge distillation. A noteworthy example of transformer distillation is DistilBERT \cite{DistilBert}, a distilled version of BERT \cite{Bert}. DistilBERT is impressive because it maintains most of the parent's performance, 97\%, while reducing the size by 40\%. In the process, it also gains a 60\% speed boost by eliminating half of the blocks, copying the weights of the rest, and using soft labels according to the probability distribution of the teacher. 

\par \textbf{Transformers}, a family of attention-centric models, represent a milestone in the evolution of deep learning. Originally designed for text, they have overtaken recurrent neural networks \cite{elman1991distributed} due to their superior context awareness \cite{Bert}. Transformers have also shown themselves capable of handling images using the ViT by matching or exceeding SOTA performance \cite{liu2021swin}.  
The idea is to consider patches of 16x16 pixels, which are embedded into standard transformer encodings and treated like word embeddings. Since images are inherently two-dimensional, unlike text, special care is given to the positional encodings so that they relay correct information about the positioning of the patch in the image. We choose ViT because a good solution to our problem must localize the skin lesion and ignore the rest, which suits the attention mechanism perfectly.

\section{Method}

\subsection{Vanilla SkinDistilViT}
\par

\par \textbf{Teacher Model.} To the best of our knowledge, there is no vanilla ViT trained for the ISIC 2019 challenge and with publicly available weights. Therefore, we train one ourselves. The ViT is one of the models supported by the Huggingface library \cite{huggingface-transformers}, a popular open-source project for experimenting with and running transformer models. Although it is also available through the vanilla PyTorch\footnote{\url{https://pytorch.org/}, last visited March 2023.}, we opt for the Huggingface library version\footnote{\url{https://huggingface.co/google/vit-base-patch16-224}, last visited March 2023.} to train our models because of its user-friendliness and training optimizations.

\par Training transformers from scratch without considerable data is a bad idea since their scale and attention mechanism make training unstable \cite{initialization-importance}. Since we have a fairly limited dataset, we rely on transfer learning from the existing ViT trained on ImageNet \cite{imagenet}. Because it is not a tiny dataset (25k images), we fine-tune all parts of the model and follow the standard training procedure, with the default hyperparameters provided by the framework. 

\par \textbf{Student  Model.} Since transformers are highly modular, we follow the example of DistilBERT and eliminate half of the encoder blocks to create the student model. Out of the 12 blocks of the original ViT, we keep only blocks 0, 2, 4, 7, 9, and 11. We perform this by altering the state dictionary of the bigger model. We keep all non-transformer block parameters. We solve the weight initialization problem by copying the weights of the selected transformer blocks of the teacher model. We also copy all the other trainable weights. 


\par \textbf{Loss Functions.} Similar to DistilBERT, we use a mix of losses for fine-tuning our distilled model. Besides the original training objective, we use a cross-entropy loss between the teacher's and the student's outputs and a cosine loss between their hidden states. We also experiment with a mean square error (MSE) loss between the logits of the two networks. These losses are combined linearly to obtain the final loss, with their weights representing training hyperparameters. 

\subsection{Full Distillation}

\par We are interested in providing models of different sizes and levels of performance. Therefore, we propose three techniques to obtain models ranging from a full configuration of twelve transformer blocks to models with only a few, even one. We call the process full distillation.

\par First, we study a ViT model that outputs a prediction for the class at every stage by adding an independent prediction head at each of them. We try two approaches. In the first case, we use the hidden states of each classification layer independently. We use the usual cross-entropy loss at every layer and combine them linearly. We call this \textbf{Full Classification ViT (FCViT)}. This approach injects gradient at each level and forces the model to find the best features for our task early. In the second case, we link the classification heads of each level by pushing  the resulting probability distribution to match the one of the next level by employing a Kullback-Leibler divergence loss \cite{kullback1951information} while keeping the cross-entropy loss only for the topmost layer. We call this \textbf{Full Classification ViT with Probabilities (FCViTProbs)}. For convergence, we employ a multi-step training approach in which we train only the final classification head for $M$ epochs, then add the classification heads one by one every $N$ epochs, starting from the last but one downwards and finishing by fine-tuning the whole stack for another $P$ epochs. This creates an implicit distillation process in the same model without separate training. Both solutions make an implicit stack of models that can be used standalone for classifying the result, all with a single training.

\par The second approach is to progressively distill the model, eliminating one transformer block at a time. The idea is to let the model concentrate the information as well as possible by eliminating minimal capacity, unlike SkinDistilViT's case, where we eliminate half the capacity from the start. We name this process \textbf{Cascading Distillation ViT}. The idea is similar to FCViTProbs, but we ensure that all the possible knowledge is kept from one layer to the next by forcing both the probability distribution matching and the correct task predictions. In this case, for a model with $k$ layers, the teacher is the model with $k+1$ layers. For the full model, we use the full FCViT as a teacher but keep the same size. The subsequent students are initialized from the previous student by stripping it of the last transformer block. We do this for the whole stack until only one transformer layer is kept. 

\section{Experimental Setup}


\subsection{Dataset}

\par The ISIC 2019 challenge proposed a hefty set of 33,569 high-quality dermoscopic images of skin lesions, classified into eight categories, as follows: 
Melanoma, Melanocytic nevus, Basal cell carcinoma, Actinic keratosis, Benign keratosis, Dermatofibroma, Vascular lesion, and Squamous cell carcinoma. 
This dataset is split into 25,331 annotated images (i.e., the training set) and 8,238 images without public annotations (i.e., the test set). The whole dataset is a combination of three corpora, namely BCN\_20000 \cite{BCN}, HAM1000 \cite{HAM10000}, and MSK \cite{ISIC_orig}. We use this dataset for training our models.

Since the official test set of ISIC 2019 is not available, we split the existing labeled data into 80\% training and 20\% test, taking class imbalance into account. 
The class imbalance problem is major, with the most populated class having more than ten thousand samples and the least populated one having only several hundred samples. 

\par \textbf{Data Augmentation.} A downside of small datasets is that they do not provide sufficient variety. This can lead to models that have a hard time generalizing. Since no two melanoma are the same, this is a dangerous shortcoming. We address this problem by employing augmentation techniques. All images of the dataset have the lesion close to the center. To avoid this bias, we use random cropping but keep the target size large enough so that the lesion is still fully present. We also apply: (i) basic spatial transformations, like shift, scale, and rotate, and (ii) color augmentations, like RGB shift and randomly changing the brightness or contrast.

\subsection{Compared Methods}

\par \textbf{CNN Models.} We train two CNN baselines. For the first baseline, we choose the EfficientNet-B4 because it stands in the middle of the EfficinetNet-B0 to B7 family and is part of the ensembles of the top two best-performing solutions from the competition. For the second baseline, we train the EfficientNet-B6 because it has roughly the same number of parameters as our SkinDistilViT. We train the EfficientNets using Pytorch Lightning\footnote{\url{https://www.pytorchlightning.ai/}, last visited March 2023.}. For fairness, we employ the same augmentations as ViT.

\par \textbf{ViT.} The official results of the ISIC 2019 challenge are incompatible with our experiments. On the one hand, they lack public labels, so we cannot compute our performance on them. On the other hand, we cannot submit our models to the competition because the official test set contains images from categories never seen at train time, which should be labeled as "unknown". This task is of no interest to our use case, so we omit it. Thus, we rely on training a baseline ViT ourselves.

\par \textbf{SkinDistilViT.} It is initialized by transferring the weights from the ViT model trained on our task. SkinDistilViT is trained with both the task loss and the cross-entropy loss, combined with weights 1 and 0.5, respectively. Unlike DistilBERT, SkinDistilViT did not benefit from adding either the hidden cosine loss or the MSE logit loss; therefore, we omit those results for brevity.

\par \textbf{SkinDistilViT Variants.} We study the importance of weight initialization by training four versions of SkinDistilViT, all starting from the same architecture but with different initializations, as follows:

\begin{itemize}
\item DistiViT\_fs \--from scratch\-- has its parameters initialized randomly, so it does not benefit from pre-training.
\item  SkinDistilViT\_fi \--from ImageNet\-- has its parameters extracted from the original, pre-trained on ImageNet, ViT.
 \item  SkinDistilViT\_nt is an untrained version whose weights are just copied from its teacher.
\item  SkinDistilViT\_t is a version trained only with the task loss.
\end{itemize}
    
\subsection{Evaluation Metrics}

\par Because of the inherent imbalance of the real data in medical scenarios, we resort to the balanced multi-class accuracy (BMA) as a metric for comparing results, as suggested in the ISIC 2019 challenge. We also employ the standard metrics for classification tasks, namely accuracy, precision, recall, and the F1-score, in their weighted form. We compute all these metrics using the official TorchMetrics implementation of PyTorch.

\subsection{Implementation Details}

\par We have done all experiments on a machine with an i5-13600K paired to a 3090 Ti with 32GB RAM. All models are trained on the same training set, with the same augmentations, for 20 epochs. The batch size is 64 for all transformer models. The EfficientNets require more memory at training time, and therefore, we use a batch size of 32 for EfficientNet-B4 and 8 for EfficientNet-B6. 

\section{Results}

\subsection{Performance Comparisons}
\par  Table \ref{performance_comparison} depicts the comparison between the models. SkinDistilViT obtains great results, the best of all SkinDistilViT variants, proving the importance of teacher guidance. All fine-tuned SkinDistilViTs beat both CNNs in all metrics. The untrained SkinDistilViT does surprisingly well, too, considering many connections from its parent have been cut. We argue that this is due to the skip connections of the transformer block.

\begin{table}[h]
    \caption{Model performance comparison. The top part compares ViT to EfficientNets, while the bottom part compares SkinDistilViT variants. Bold indicates the best score for each metric, per comparison.}
    \label{performance_comparison}
    \centering
    \begin{tabular}{c|c|c|c|c|c}
        \toprule
            Model & BMA (\%) & Accuracy (\%) & Precision (\%) & Recall (\%) & F1-score (\%) \\
        \midrule
            EfficientNet-B4            & 27.64             & 71.60     & 67.95     & 71.62  & 67.80  \\
            EfficientNet-B6            & 31.51             & 81.62    & 86.25     & 81.62  & 82.21 \\
            ViT  & \textbf{83.73} & \textbf{89.18} & \textbf{89.04}  & \textbf{89.18}  & \textbf{89.06} \\
            \midrule
            SkinDistilViT\_nt & 42.00             & 66.60     & 64.25     & 66.60   & 64.24 \\
            SkinDistilViT\_fs & 39.88             & 66.79    & 63.67     & 66.79  & 64.03 \\
            SkinDistilViT\_fi & 80.23             & 86.68    & 86.49     & 86.68  & 86.51 \\
            SkinDistilViT\_t  & 80.96             & 87.80     & 87.60      & 87.80   & 87.61 \\
            SkinDistilViT & \textbf{82.34} & \textbf{88.51} & \textbf{88.34} & \textbf{88.51}  & \textbf{88.37} \\
        \bottomrule
    \end{tabular}
\end{table}

\par The behavior of the ViTs in the case of imbalanced classes is interesting when compared to CNNs. While the CNNs are greatly affected by the imbalance, as seen in the BMA score, the ViTs seem unfazed. We suspect this stems from the filter-based nature of the CNNs, which makes them more reliant on the image's texture. 

\par
\begin{figure}[th]
    \centering
    \includegraphics[width=0.8\textwidth]{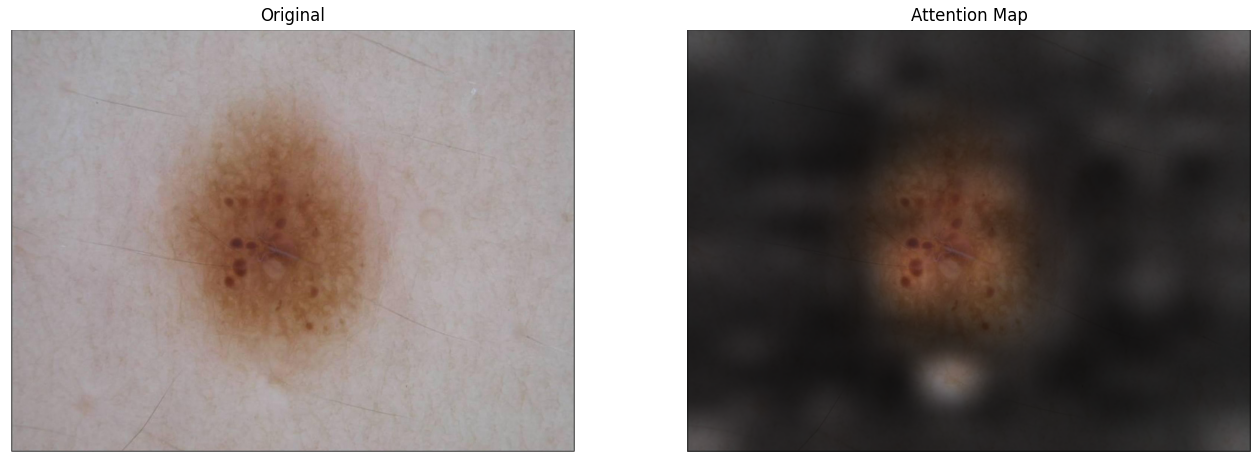}
    \caption{Attention map visualization with BertViz \cite{vig-2019-multiscale}: original image without and with the attention map applied.}
    \label{fig:attention}
\end{figure}

 Since all images contain skin, they can get more easily confused. The better behavior of ViT can be explained by the attention mechanism, which only ensures the processing of the relevant part of the image. The attention mechanism in action can be observed in Figure \ref{fig:attention}, where the model only pays attention to the skin lesion. The class imbalance problem, although alleviated, is still present because the BMA score is several points lower than all the others.

\subsection{Full Distillation Results}


\par We train the three full distillation experiments. The results of all three approaches can be observed in Table \ref{Full Distillation}. FCViT and FCViTProbs are trained starting from the original ViT trained on ImageNet, just like our ViT baseline. For the cascading distillation process, we use as a teacher the best-performing full model we had obtained that far, FCViTProbs, which has a BMA of 85.13\% compared to the 83.73\% BMA of the base ViT. We do not start from the same model because we are more interested in the loss of performance rather than actual numbers. 

\begin{table}[h]
\caption{Full distillation results in terms of BMA. Lx represents the classification layer we computed the result from. Bold indicates the best score for each line.}
\label{Full Distillation}
\centering
\begin{tabular}{c|c|c|c|c|c}
    \toprule
 Last Layer & \begin{tabular}[c]{@{}c@{}} FCViT \\(\%) \end{tabular}& \begin{tabular}[c]{@{}c@{}}  FCViTProbs \\ (\%) \end{tabular}& \begin{tabular}[c]{@{}c@{}} Cascading Distillation\\ ViT (\%)\end{tabular} & \begin{tabular}[c]{@{}c@{}} ViT \\(\%) \end{tabular}& \begin{tabular}[c]{@{}c@{}} SkinDistilViT \\(\%)\end{tabular} \\
    \midrule
L0                                                             & 31.23 & 12.26      & \textbf{42.86}                                                            &    -   &     -      \\
L1                                                             & 50.42 & 13.69      & \textbf{65.18}                                                            &   -    &    -       \\
L2                                                             & 63.53 & 12.30       & \textbf{75.71}                                                            &    -   &     -      \\
L3                                                             & 74.19 & 17.70       & \textbf{80.90}                                                             &   -    &     -      \\
L4                                                             & 79.16 & 27.72      & \textbf{81.75}                                                            &   -    &     -      \\
L5                                                             & 82.27 & 40.52      & \textbf{83.39}                                                            &   -    & 82.34     \\
L6                                                             & 83.46 & 52.77      & \textbf{84.33}                                                            &   -    &     -      \\
L7                                                             & 84.27 & 61.26      & \textbf{84.74}                                                            &   -    &     -      \\
L8                                                             & 84.20  & 66.18      & \textbf{85.16}                                                            &   -    &     -      \\
L9                                                             & 84.57 & 73.70      & \textbf{85.16}                                                            &   -    &    -       \\
L10                                                            & 84.66 & 82.50      & \textbf{85.54}                                                            &   -    &      -     \\
L11                                                            & 84.68 & 85.13      & \textbf{85.83}                                                            & 83.73 &     -     \\
    \bottomrule
\end{tabular}
\end{table}


\par Training everything on one go behaves well in the FCViT case. Thus, it is stable, and its performance is more than adequate. This approach matches the SkinDistilViT in terms of performance, without extra training and guidance from a teacher model, while surpassing the SkinDistilViT\_t considerably. 
However, at lower dimensions, the performance greatly diminishes. Thus, the cascading distillation approach is more suitable for tiny models. The probability distribution matching ViT behaves rather badly, not managing to give good classifiers, especially at lower levels. However, the approach seems to help with training the network because the performance of the full model surpasses both the ViT and the FCViT, respectively.


\par Cascading Distillation ViT obtains the best results at all levels, but it shines in preserving the performance at a lower number of layers. All full-size models surpass the original ViT model. The original SkinDistilViT is still competitive, behaving slightly better than the similarly sized FCViT.

\subsection{Distillation Trade-off}

\par A trade-off analysis between SkinDistilViT and ViT can be observed in Table \ref{trade-off}. In general, the performance loss is low, while the gains are considerable. The loss is greater in BMA's case, which indicates that the smaller model loses more nuances. 

\par Regarding speed, we run the same scenario for both CPU and GPU. We measure the speed by dividing the number of test samples by the inference time of the model, ignoring batching and data loading. 
Interestingly, the speed gain on the CPU is larger. This is explainable by the differences in the design of the two processing units. Thus, GPUs are designed for matrix multiplications and deal with great deals of data in parallel, so the speed does not double. Instead, CPUs are more general, which translates into the expected double speed.

\begin{table}[h]
    \caption{Distillation trade-off.}
    \label{trade-off}
    \centering
    \begin{tabular}{c|c|c|c|c|c}
        \toprule
        & \begin{tabular}[c]{@{}c@{}} BMA \\ (\%) \end{tabular} & \begin{tabular}[c]{@{}c@{}} Recall \\ (\%) \end{tabular} & \begin{tabular}[c]{@{}c@{}} Speed CPU \\ (it/s) \end{tabular}& \begin{tabular}[c]{@{}c@{}} Speed GPU \\ (it/s) \end{tabular}& \begin{tabular}[c]{@{}c@{}} \#Params \\ (Millions) \end{tabular} \\    
        \midrule
    ViT       & 83.73     & 89.10  & 10.79            & 206.31           & 85.85              \\
    SkinDistilViT & 82.43     & 88.51  & 21.36            & 349.20            & 43.27              \\
        \midrule
    Gain      & -1.57\%   & -0.60\%  & 97.96\%          & 69.25\%          & 49.60\%            \\
        \bottomrule
    \end{tabular}
\end{table}

\par  Regarding the speed comparison between ViT and CNNs, EfficientNet-B6 has a speed of 64.67 items/second on GPU, while the similarly sized SkinDistilViT sits at 349.2 items/second, 5.4 times faster. Training the SkinDistilViT took 81 minutes for 20 epochs, while training the EfficientNet-B6 took 364 minutes for the same number of epochs, 4.49 times more. The convolution operation explains the difference because it uses the same parameters for many operations. This means the EfficientNet does more operations than the ViT for the same number of parameters, hence the lower speed.

\par
\begin{figure}[th]
    \centering
    \includegraphics[width=0.7\textwidth]{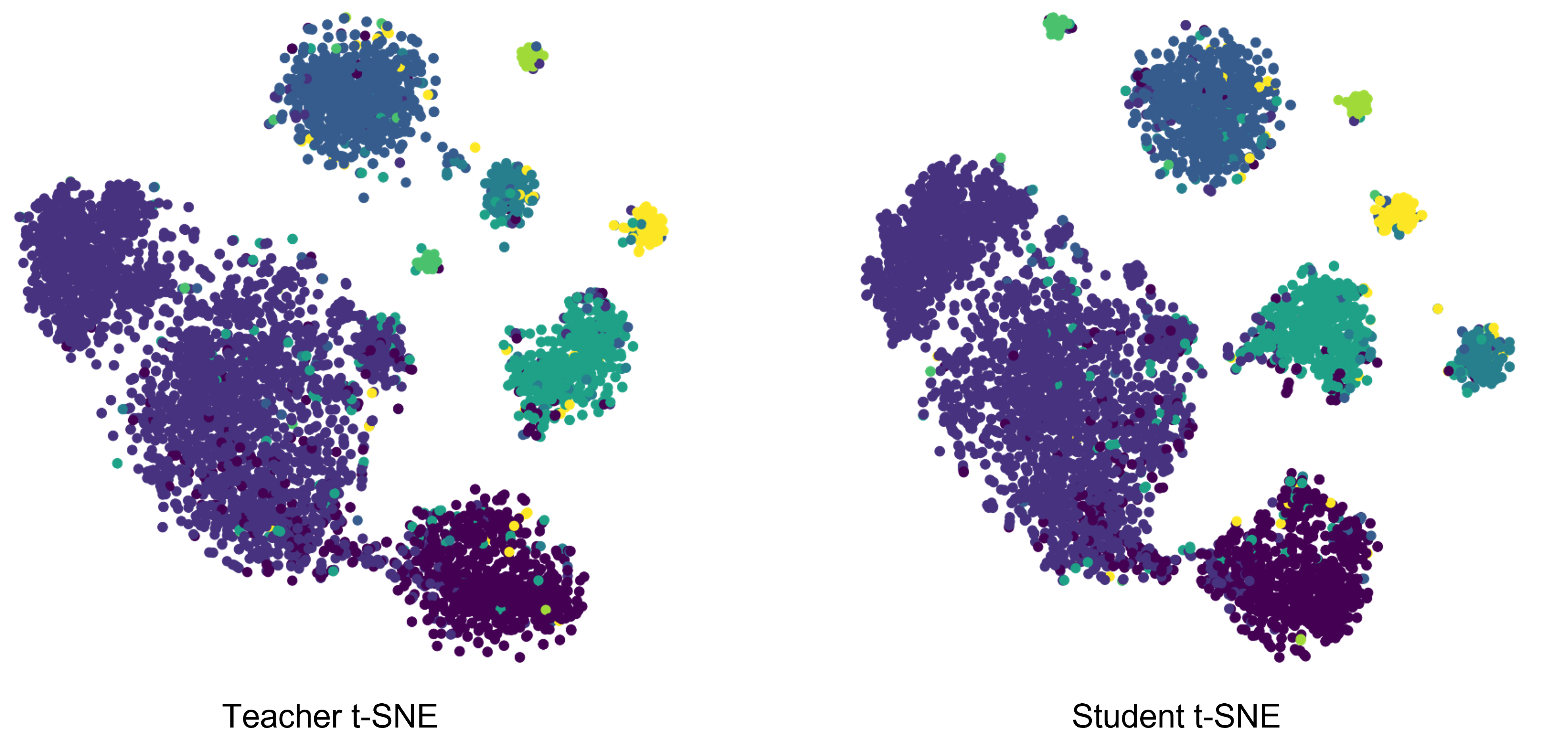}
    \caption{Visualizations of two-dimensional t-SNE embeddings for the teacher (i.e., ViT) and student (i.e., SkinDistilViT) models. The student embeddings are noisier but keep the separation of the teacher classes well.}
    \label{fig:tSNE}
\end{figure}

 A comparison of the expressiveness between the embeddings of the teacher and the student can be observed in Figure \ref{fig:tSNE}. We use the t-distributed
stochastic neighbor embedding (t-SNE) technique \cite{tSNE} to project the high-dimensional embedding provided by the ViT to a two-dimensional space. The teacher model separates the eight classes well, with clearly defined clusters, regardless of the class imbalance. The student model, albeit noisier, keeps the same performance.

\subsection{Cancer Detection Performance}

\par
\begin{figure}[th]
    \centering
    \includegraphics[width=0.6\textwidth]{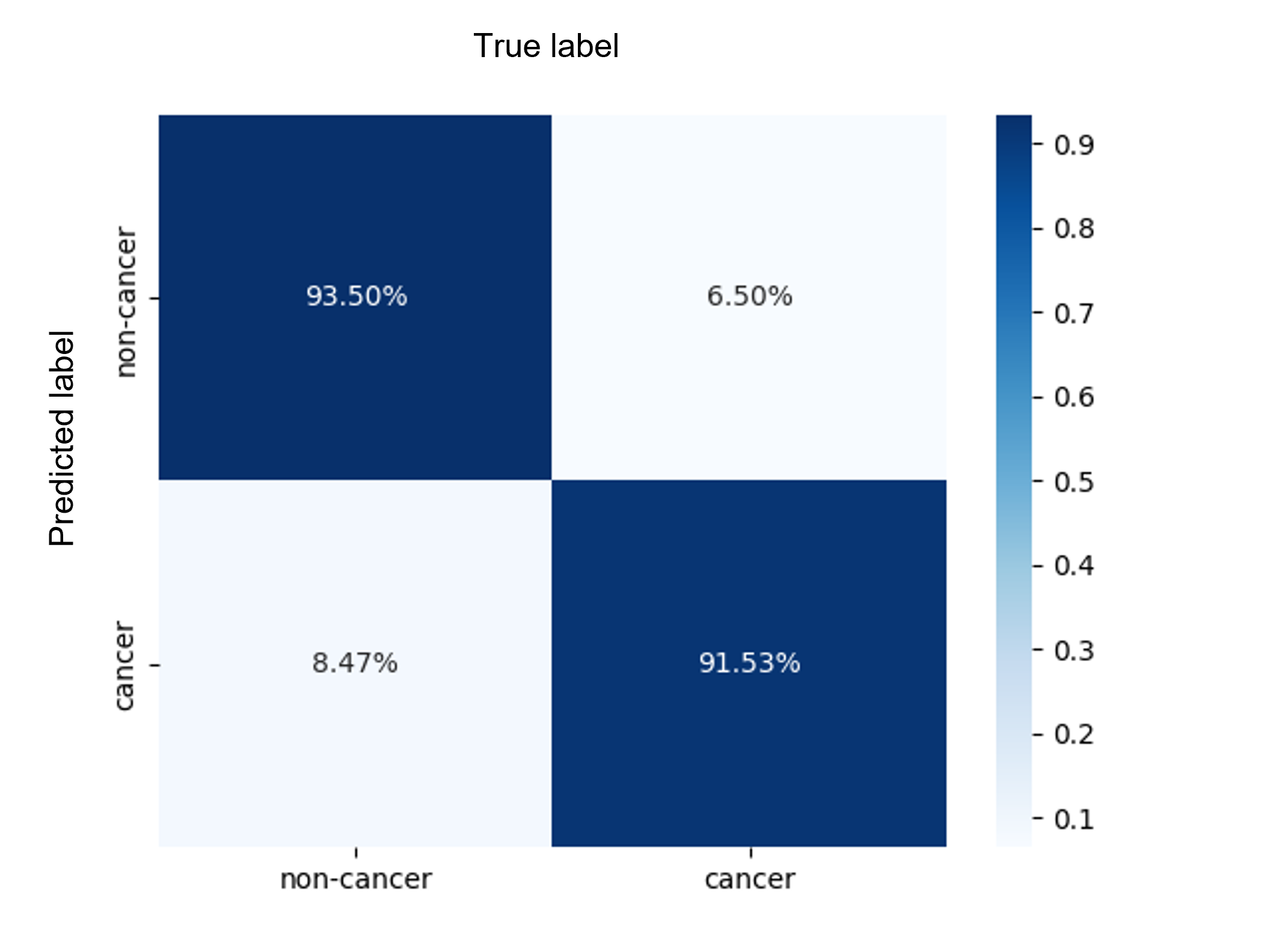}
    \caption{The confusion matrix for our Cascading Distillation ViT on cancer versus non-cancer problem.}
    \label{fig:cancer_vs_noncancer}
\end{figure}

 Another performance metric is how well the model determines whether a lesion is cancer or not. This metric is not directly computed through the dataset. We compute it by separating the classes into cancer classes (i.e., Melanoma, Basal cell carcinoma, and Squamous cell carcinoma) and benign classes (i.e., Melanocytic nevus, Actinic keratosis, Benign keratosis, Dermatofibroma, and Vascular lesion). When analyzing the results based on this split, we obtain an accuracy of 92.8\%, a precision of 91.53\%, a recall of 86.73\%, and an F1-score of 89.06\%. The confusion matrix for this problem can be found in Figure \ref{fig:cancer_vs_noncancer}.

\par
\begin{figure}[th]
    \centering
    \includegraphics[width=0.9\textwidth]{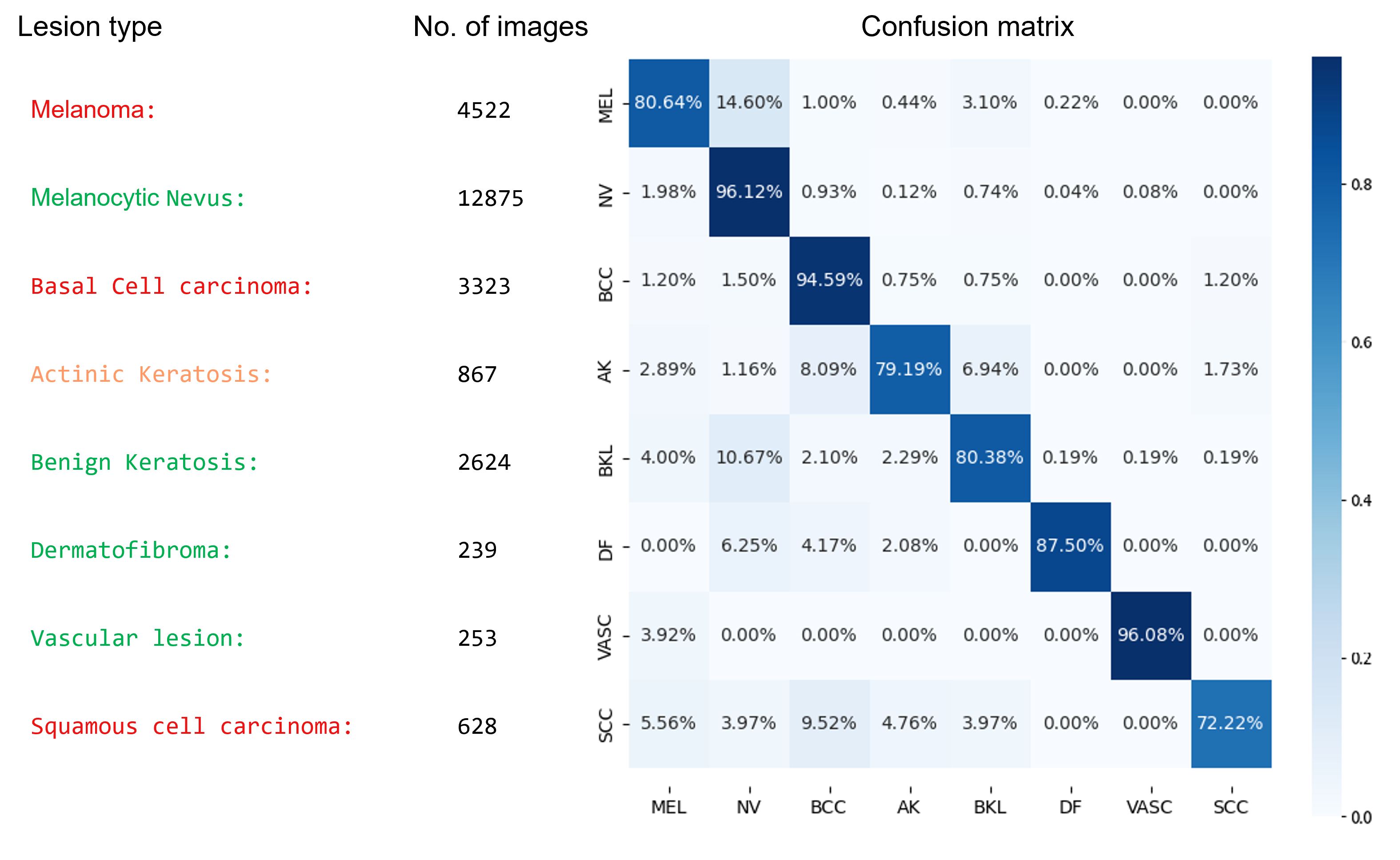}
    \caption{The confusion matrix of the Cascading Distillation ViT (full size, 12 layers). Red represents malignant classes, green represents benign classes, and orange represents benign classes that can turn malignant.}
    \label{fig:conf_matrx}
\end{figure}

 Our best-performing model classifies melanoma correctly in 80.64\% of the cases. The human baseline for this operation, i.e., dermatologists with ten years of experience, is 80\% \cite{melanoma_baseline}, which means we obtain good performance.
The confusion matrix of the classification results is given in Figure \ref{fig:conf_matrx}.
The confusions of our model are the ones we would expect.
The lesions are often confused with common moles (Melanocytic nevus), which is a common mistake. Also, Actinic keratosis is confused with Benign keratosis, another type of keratosis. Moreover, Squamous cell carcinoma is confused with Basal cell carcinoma, another type of cancer. Interestingly, the class imbalance is not necessarily a problem since the top scores are not obtained in the highest populated classes, nor the worst scores in the least populated ones. 

\section{Conclusions and Future Work}
\par In this work, we provided a model for the skin lesion classification that is lightweight, yet performant. The resulting distilled network is strong, keeping most of the performance while considerably increasing speed and decreasing memory consumption. Due to the attention mechanism, it has also proven superior to CNNs in terms of performance, especially considering class imbalance.

\par Careful weight initialization is critical to a good model. Training a distilled model from scratch provides worse results than simply copying weights from the bigger model. ImageNet initialization is good, but starting from the fine-tuned ViT is better. Teacher guidance completes the distillation by providing a good performance increase.

\par By forcing a consistency loss between layers and employing cascading distillation on top of the resulting model, we were able to boost the performance across all numbers of layers compared with the standard ViT and SkinDistilViT. Moreover, this technique creates a family of well-performing models of different sizes.

\par Last, our full-size models surpass the human baseline on melanoma identification and almost match it in the distilled form, while providing solid results for the skin cancer identification problem.

\par As future work, we propose combining the three full distillation techniques. We can add all the classification heads and train them as in FCViT but also force the probability distributions of their outputs to match, then apply the cascading distillation process. Another next step would be to study the impact of class imbalance. Although robust to it, SkinDistilViT might benefit from a balanced dataset. 

\section*{Acknowledgments}
This research has been funded by the University Politehnica of Bucharest through the PubArt program.

\bibliographystyle{splncs04}
\bibliography{samplepaper}

\end{document}